\documentclass[runningheads]{llncs}
\usepackage[T1]{fontenc}
\usepackage{url}
\usepackage{hyperref}
\usepackage{graphicx}
\graphicspath{{./}}
\usepackage{amsmath}
\usepackage{booktabs}
\usepackage{algorithm}
\usepackage{algorithmic}
\usepackage[switch]{lineno}
\usepackage{tabularx}
\usepackage{changes}
\usepackage{enumitem}
\usepackage{multirow}
\usepackage{pifont}
\newcommand{\BibTeX}{B\kern-.05em{\sc i\kern-.025em b}\kern-.08em\TeX}
\newcolumntype{M}[1]{>{\centering\arraybackslash}m{#1}}
\newcolumntype{P}[1]{>{\centering\arraybackslash}p{#1}}
\newcommand\presto{\textsf{Presto}}

\DeclareMathOperator*{\argmax}{arg\,max}

\begin{document}
\title{presto: Efficient, Training-free, and Open-world Object Placement via Imaginary Search}
\titlerunning{presto: Efficient, Training-free, and Open-world Object Placement}
%
\author{Weixuan Ding\inst{1} \and
Shang Liu\inst{2} \and
Hanyu Pei\inst{2} \and
Zeyan Liu\inst{2,}\thanks{Corresponding author.}}
\authorrunning{W. Ding et al.}
%
\institute{Wuhan University, Wuhan Hubei Province, China \\
\email{weixuan.ding@outlook.com} \and
University of Louisville, Louisville KY, USA \\
\email{\{shang.liu, hanyu.pei, zeyan.liu\}@louisville.edu}\\
}
\maketitle              
\begin{abstract}
Object placement is critical in image composition, requiring spatially and semantically coherent positioning of objects within diverse scenes. Existing approaches typically rely on hand-crafted rules or supervised learning on limited datasets, which restricts their generalization and interpretability, especially in open-world scenarios involving novel objects and scenes. In this work, we reformulate open-world object placement as a heuristic search task guided by reasoning from a Multimodal Large Language Model (MLLM). We introduce \textsf{presto}, a zero-shot, training-free framework that operates within an imaginary action space to iteratively refine object position and scale. Our coarse-to-fine search strategy ensures fast convergence, and we evaluate two decision-making variants: Metric-guided Selection and MLLM-as-a-judge. Experiments across multiple benchmarks show that \textsf{presto}~achieves state-of-the-art performance, particularly in previously unseen, open-world settings. Human studies further reveal that the MLLM-as-a-judge variant produces more perceptually coherent placements than metric-driven approaches, highlighting a gap between standard evaluation metrics and human visual judgment.

\end{abstract}

\section{Introduction}
Image composition involves synthesizing a composite image by inserting a foreground object into a background scene~\cite{niu2025making}. This task underpins a wide range of applications, including image editing~\cite{chen2019realistic}, augmented reality~\cite{lee2018context}, artistic design~\cite{zhang2023interactive}, and synthetic data generation for training vision models~\cite{ouyang2018pedestrian,Song_2024}. A central challenge is resolving inconsistencies between the inserted object and the background in terms of boundary alignment, appearance, geometry, occlusion, and semantic coherence~\cite{niu2025making}. While diffusion-based generative models have made substantial progress in visual blending~\cite{zhang2023controlcom,wang2024primecomposer,chen2024anydoor,wang2025unicombine}, the \textit{object placement} problem, which is to determine spatially and semantically plausible positions and scales, remains underexplored.

Existing object placement methods can be broadly categorized into two groups. Early rule-based approaches rely on handcrafted constraints to determine object position and scale~\cite{remez2018learning,georgakis2017synthesizing,zhang2018unreasonable,fang2019instaboost}, but often result in implausible or inconsistent placements. More recent supervised methods train end-to-end models on annotated datasets~\cite{tripathi2019learning,zhang2020learninga,zhou2022learning,wang2023cagan,wang2024learning,zhang2023interactive,qin2025think}. While these methods show promise in controlled settings, they suffer from several key limitations:
(1) High training cost: Supervised approaches require significant resources, including time-intensive annotation of positive and negative placements, and considerable computational overhead for training and tuning.
(2) Limited dataset coverage: Datasets such as OPA~\cite{liu2021opa} cover only a narrow set of object categories (e.g., 47 types) and may include suboptimal or biased annotations, which limits generalization to open-world scenarios.
(3) Lack of interpretability: These models typically offer little insight into why certain placements are preferred, making their decisions difficult to understand or trust.

In contrast, real-world object placement demands \textit{open-world generalization}, which is the ability to position novel objects in unseen scenes without task-specific training. This requires both spatial reasoning (e.g., ensuring physical support and valid occlusion) and semantic understanding (e.g., placing a cake on a table, not on a keyboard). The task also involves a continuous solution space, where many valid placements may exist simultaneously.

Multimodal Large Language Models (MLLMs), such as GPT-4o~\cite{hurst2024gpt4o}, offer strong potential for this task. They combine visual perception with common-sense reasoning and have demonstrated success across diverse vision-language tasks, including object detection~\cite{tang2023cotdet,wu2024dettoolchain}, embodied AI~\cite{ni2024cotdiffusion,dream2real}, and multimodal generation~\cite{zhou2024minedreamer}. In particular, Chain-of-Thought (CoT) prompting has further enhanced their ability to reason through complex problems.

Motivated by these strengths, we investigate whether prompting-based MLLMs can be applied to open-world object placement. We find that although these models exhibit strong general reasoning, they struggle with this task, especially when dealing with separate foreground and background inputs. To address this, we reformulate object placement as a heuristic search problem guided by MLLM reasoning, rather than a one-shot prediction. 

We instantiate this idea with \presto, a training-free, zero-shot framework for open-world object placement. \presto~generates multiple initial placement candidates and iteratively refines their position and scale within an imaginary action space—a defined set of possible transformations—based on step-by-step feedback from the MLLM. A multiscale search strategy enables rapid convergence by starting with coarse adjustments and progressively narrowing the search space. We also introduce two selection strategies: \ding{182} Metric-guided Selection, which optimizes an existing evaluation metric, and \ding{183} MLLM-as-a-judge, which prompts the MLLM itself to assess placement plausibility. Experiments on several benchmarks show that \presto~achieves state-of-the-art performance, outperforming supervised baselines. It especially excels in novel, open-world scenarios, where existing methods often fail to generalize. Notably, although MLLM-as-a-judge scores lower on automatic metrics, it performs better in human evaluations, revealing a disconnect between current metrics and human perceptual judgment.

Our contributions are as follows:
\begin{itemize}
    \item We are the first to reformulate open-world object placement as a heuristic search problem guided by MLLM reasoning, enabling iterative spatial refinement beyond one-shot predictions.
    \item We present \presto, a lightweight, training-free, and generalizable framework that leverages MLLM-guided actions to optimize object position and scale, achieving strong performance across both benchmarks and human evaluations.
    \item We compare two selection strategies and find that MLLM-based judgments align more closely with human preferences, suggesting a gap between current metrics and perceptual quality.
\end{itemize}

\begin{figure*}[t]
  \centering
  \includegraphics[width=\linewidth]{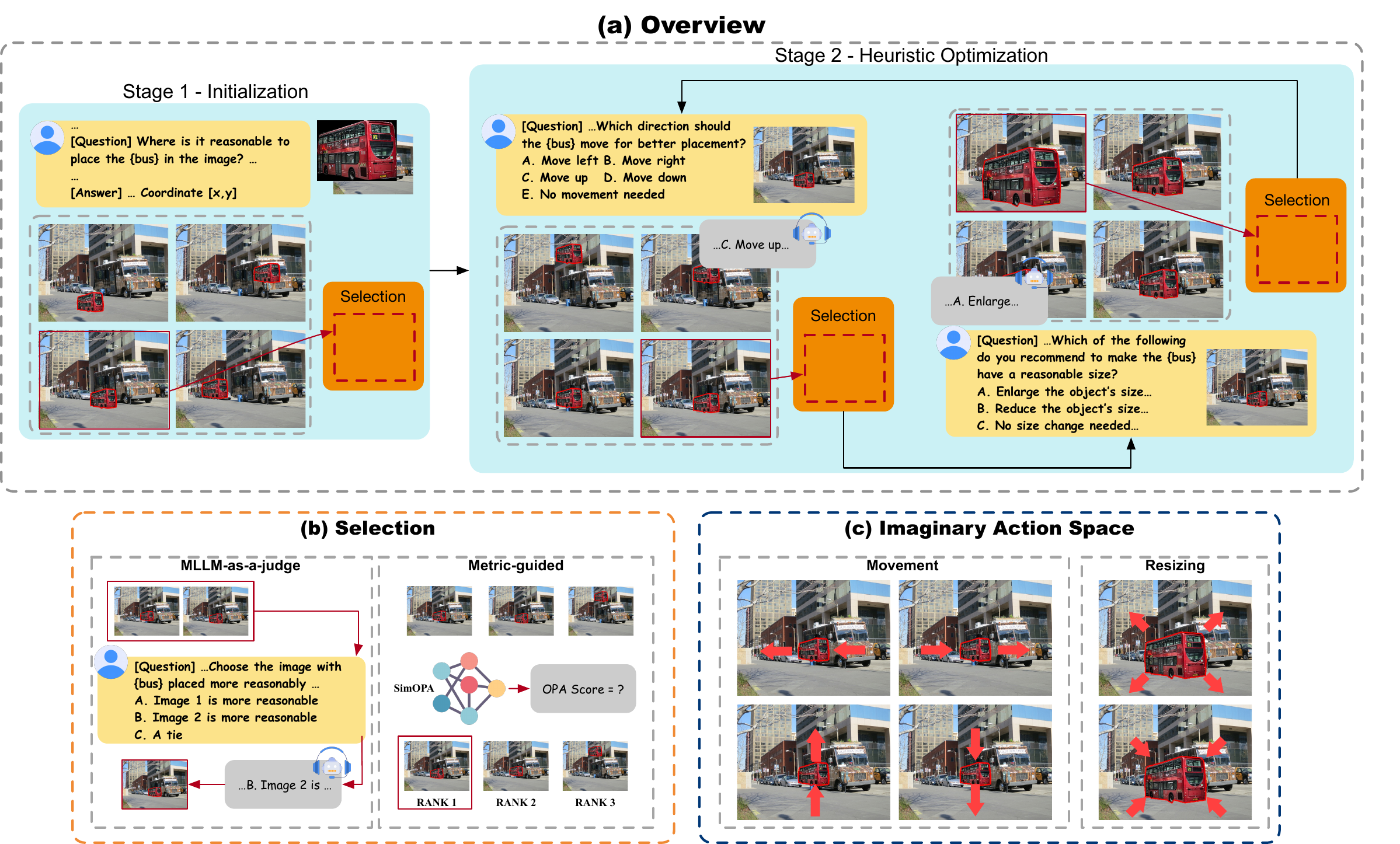}
  \caption{Visualization of qualitative comparison of the proposed method and the baselines on OPA dataset. Foreground is outlined in red}
  \label{fig:method}
  \vspace{-0.5cm}
\end{figure*}

\vspace{-0.4cm}
\section{Related Works}
\textbf{Object placement} is a computer vision task that involves inserting a foreground object into a background image in a way that looks natural and contextually appropriate~\cite{niu2025making}. A reasonable placement requires carefully choosing the object’s location, size, and shape so that it fits seamlessly into the scene. This helps avoid unrealistic outcomes like disproportionately sized objects, awkward occlusions of background, or physically impossible positioning. Traditional approaches are mostly rule-based, relying on geometric and depth constraints~\cite{remez2018learning,georgakis2017synthesizing}, class-specific consistency~\cite{zhang2018unreasonable}, or appearance consistency heatmaps~\cite{fang2019instaboost}. However, these methods often lack precision and practical usability. More recent techniques use deep neural networks, which can be categorized into category-specific and instance-specific methods \cite{niu2025making}. Category-specific methods \cite{tan2018where,lee2018context,dvornik2018modeling,dvornik2019importance,volokitin2020efficiently} predict plausible bounding boxes for object categories, providing a generalized solution for all instances within the same category. Instance-specific methods \cite{tripathi2019learning,zhang2020learninga,zhou2022learning,wang2023cagan,wang2024learning,zhang2023interactive,qin2025think} determine the best transformations for each unique pair of foreground and background images. 
Notably, a key evaluation tool in this line of work is SimOPA~\cite{liu2021opa}, a binary classifier trained to distinguish rational and irrational composites. It is widely used as the benchmark for assessing placement quality~\cite{zhou2022learning,wang2024learning,wang2023cagan,zhang2023interactive,qin2025think}.

\noindent\textbf{Multimodal Large Language Models (MLLMs)} are designed to process and reason over inputs from multiple modalities.
Leading models typically use a unified encoder-decoder architecture, where multimodal inputs are first transformed into a shared embedding space and then decoded to generate coherent text outputs. This area has advanced rapidly, especially with the rise of commercial models like GPT-4V~\cite{achiam2023gpt4}, Gemini 2.0~\cite{team2023gemini}, and Claude 3~\cite{Anthropic2024claude3}. Vision-Language Models (VLMs) such as BLIP-2~\cite{li2023blip}, OpenFlamingo~\cite{awadalla2023openflamingo}, MiniGPT-4~\cite{zhuminigpt}, and LLaVA~\cite{liu2023llava,liu2024llava1.5,li2024llavaonevision} are trained on paired image and text data to align visual and textual understanding. Models such as VideoChat~\cite{li2023videochat} and Video-ChatGPT~\cite{maaz2023video} extend these capabilities to video understanding.
Recent progress in MLLMs has focused on improving reasoning capabilities. 

\noindent\textbf{Chain-of-Thought (CoT) reasoning}~\cite{wei2022CoT} encourages large language models to solve problems step by step, improving performance on complex tasks and making decision-making transparent. While traditional CoT follows a linear, sequential chain, newer approaches explore more flexible structures like trees~\cite{long2023ToT,yao2023ToT} and graphs~\cite{besta2024GoT}. These developments have led to more dynamic and interpretable rationales in multimodal reasoning~\cite{wang2025multimodal}.

\vspace{-0.4cm}
\section{Preliminaries}

\subsection{Problem Definition}
Given a background image \( \mathcal{I}_b \), a foreground image \( \mathcal{I}_f \), and a mask \( \mathcal{M}_f \) outlining the foreground object, the object placement process generates a composite image $\mathcal{I}_c$ along with its corresponding composite mask \( \mathcal{M}_c \). The composition is achieved through a placement function $F$. Following the formulation in \cite{zhou2022learning,zhang2023interactive}, this task is simplified to solving the optimal placement parameters $\xi=(x, y, r)$. Specifically, \( x \) and \( y \) specify the position coordinates of the foreground object within the background. \( r \) is the scaling factor that resizes the foreground object. This process can be formulated as:

\vspace{-0.2cm}
\begin{equation}
\langle \mathcal{I}_c, \mathcal{M}_c \rangle = F_{{x,y,r}}(\mathcal{I}_b, \mathcal{I}_f, \mathcal{M}_f)
\end{equation}

We can frame object placement as an optimization problem. Our objective is to find the placement parameters \((x^*, y^*, r^*) \) that maximize a quality score $Q$ (e.g., measured by SimOPA~\cite{liu2021opa}). $Q$ assesses how well the composition aligns visually and contextually with the original scene. This optimization problem can be formulated as:

\vspace{-0.2cm}
\begin{equation}
\label{eq:3}
\xi^*=\langle x^*, y^*, r^* \rangle = \argmax_{{x,y,r}}Q\big(F_{{x,y,r}}(\mathcal{I}_b, \mathcal{I}_f, \mathcal{M}_f)\big)
\end{equation}

While most state-of-the-art methods use specially trained end-to-end models to predict \( (x, y, r) \), we aim to leverage the general knowledge and reasoning capabilities of MLLMs, enabling a more practical, training-free alternative. 

Specifically, we adopt an MLLM which takes as input both images and a textual instruction $P$ and returns a textual response. For object placement, the image inputs include the foreground image $\mathcal{I}_f$, the foreground mask $\mathcal{M}_f$, and the background image $\mathcal{I}_b$. The instruction $P$ describes the task, provides context, and encourages reasoning. A successful response should output the predicted placement parameters \( (\hat{x}, \hat{y}, \hat{r}) \). This process can be formally defined as:

\vspace{-0.2cm}
\begin{equation}
\label{eq:5}
\hat{\xi}=\langle \hat{x}, \hat{y}, \hat{r} \rangle = \textsf{MLLM}(\mathcal{I}_b, \mathcal{I}_f, \mathcal{M}_f, P)
\end{equation}

In this setup, the optimization is performed implicitly by the MLLM through its internal reasoning. The quality of the prompt $P$ plays a crucial role in guiding the model’s output and determining the final composite image $\mathcal{I}_c$. The full prompt design used in our method is detailed in Appendix~E.

\subsection{Key Observations and Motivation}

\begin{table}[t]
    \centering
    \small
    \caption{Comparison of different MLLM reasoning techniques on 100 foreground-background pairs randomly selected from the OPA dataset.}
    \label{tab:opa}
    \begin{tabular}{p{40mm}<{\centering}|p{40mm}<{\centering}}
        \hline
        Method 
        & Accuracy\(\uparrow\) \\
        \hline
        \multicolumn{2}{c}{\textit{Zero-shot Prompting}} \\
        \hline
        GPT-4o & 71.0 \\
        DeepSeek-VL2 & 63.0 \\
        o4-mini & 65.0 \\
        \hline
        \multicolumn{2}{c}{\textit{Chain-of-Thought Prompting}} \\
        \hline
        GPT-4o+Zero-shot CoT & 78.0 \\
        GPT-4o+CCoT & 74.0 \\
        GPT-4o+Visual Sketchpad  &  44.0  \\
        \hline
        \multicolumn{2}{c}{\textit{Our Method, \presto}} \\
        \hline
        GPT-4o+\presto-Metric  & \textbf{99.0}  \\
        GPT-4o+\presto-MLLM &  \textbf{95.0} \\
        \hline
    \end{tabular}
    \vspace{-0.5cm}
\end{table}

MLLMs have recently demonstrated strong performance across a variety of vision-language tasks, including visual grounding~\cite{zhang2023llavagrounding,wu2024deepseekvl2}, object recognition~\cite{zhou2024imageofthought}, and object detection~\cite{tang2023cotdet,wu2024dettoolchain}. In particular, MLLMs have shown the ability to accurately localize objects and regions within images~\cite{xiao2024grounding,li2021GLIP}. These capabilities suggest that MLLMs possess a strong understanding of both visual and textual information, making them well-suited for addressing the challenges of object placement.

Using prompting-based methods with MLLMs instead of training dedicated predictors provides important advantages: (1) \textbf{Resource Efficiency:} Prompting does not require training or fine-tuning, making it lightweight in terms of memory and devices. It can be deployed using cloud APIs or even chatbots. (2) \textbf{Time Efficiency:} Without the overhead of model training, prompting-based workflows can be executed immediately. (3) \textbf{Data Efficiency:} Prompting does not require curated training datasets, which is particularly beneficial for object placement tasks where collecting accurate annotations is difficult and expensive. (4) \textbf{Open-world Generalization:} MLLMs can handle novel object-background pairs, not limited by predefined categories. (5) \textbf{Interpretability:} The reasoning behind decisions is transparent, clearly outlined step-by-step in the MLLM’s responses.

Motivated by these advantages, we first evaluated several advanced models and prompting strategies on the object placement task using the OPA dataset~\cite{liu2021opa}. We tested the zero-shot performance of GPT-4o~\cite{hurst2024gpt4o}, o4-mini~\cite{openai2025o3}, and DeepSeek-VL2~\cite{wu2024deepseekvl2}, the latter being specifically designed for visual grounding. Among these, GPT-4o achieved the highest accuracy at 71\%, where accuracy is defined as the percentage of placements classified as “rational” by SimOPA. We then applied reasoning-augmented prompting methods to GPT-4o, including Zero-shot CoT~\cite{zeroshot}, CCoT~\cite{mitra2024compositional}, and Visual SketchPad~\cite{hu2024visual}. However, none of these approaches produced consistently accurate placements or reliable rationales. As shown in Table \ref{tab:opa}, the overall accuracy remained low. The best result was 78\% (Zero-shot CoT), while the worst dropped to 44\%.

We attribute this pitfall to several fundamental differences between object placement and visual grounding: (1) Visual grounding operates on a single image, while object placement requires joint reasoning over two separate inputs. (2) Unlike grounding, which localizes existing content, object placement involves generating new content that must be contextually and spatially coherent with the background. (3) The object placement task lacks discrete ground-truth labels, and plausible placements vary continuously depending on position, scale, and semantic fit.

To address these challenges, we propose approaching object placement as a \textit{heuristic search} guided by MLLM reasoning. Rather than attempting to determine the optimal placement in a single step, we decompose the task into a sequence of incremental decisions. At each step, the model evaluates potential adjustments (e.g., position, size) and selects the most promising one. In this way, the placement is gradually refined and optimized. This iterative process mirrors how human beings approach similar tasks through successive thinking, adjustment, and feedback.


\vspace{-0.4cm}
\section{The~\presto~Framework}

\subsection{Overview}
Our proposed framework, \presto, integrates MLLM-driven decision-making into a local search algorithm similar to hill climbing. \presto~has three main stages: initialization, iterative heuristic optimization, and selection.

In the initialization stage, the MLLM generates multiple initial candidate placements. These candidates are evaluated for semantic quality, and the one with the highest coherence, denoted $\mathcal{I}_{c}^{(0)}$, serves as the starting point for optimization. The placement is described by its initial parameters $\xi^{(0)}=(x^{(0)},y^{(0)},r^{(0)})$.

Following initialization, \presto~iteratively refines the placement through two separate adjustments: (1) \textit{movement}, which changes the object's position $(x,y)$, and (2) \textit{resizing}, which alters the object's size $r$. These steps are conducted separately because MLLMs struggle to optimize position and size simultaneously. Our empirical observations also indicate that the position impacts semantic rationality more significantly than the size, which aligns with common sense. Thus, \presto~prioritizes movement before resizing in each iteration.

To ensure computational efficiency, our optimization is designed to converge in just a few iterations. We employ a \textit{multiscale search strategy} that begins with large movement and resizing steps to broadly explore the solution space, and gradually reduces step sizes using a decay factor $\alpha$. This approach resembles learning rate schedules in deep learning and enables a smooth transition from coarse exploration to fine-grained refinement.

\presto~also uses an elitism mechanism that consistently retains the best solution. In each iteration, it explores candidate placements through controlled movement and resizing actions. The best candidate is selected for the next iteration to continue the search. This process repeats until convergence or an early stopping criterion is met, e.g., when no improvements are found over several iterations. An overview of the framework is shown in Figure \ref{fig:method}, and the full optimization procedure is described in Appendix C.

\subsection{Imaginary Action Space}
We formally define an \textit{imaginary action space} characterized by placement parameters $(x, y, r)$. The foreground object's actual position is normalized to coordinates $(x, y) \in[0, 1]$, which represents its horizontal and vertical distances proportionally from the top-left corner of the background image. The scaling factor $r \in (0.0, 1.0]$ specifies the object's size ratio relative to the background dimensions.

Let $w_b$ and $h_b$ denote the width and height of the background image, and $w_f$ and $h_f$ denote those of the foreground object. The resized dimensions of the foreground object, denoted by $(w_s, h_s)$, are computed as:

\vspace{-0.2cm}
\begin{equation}
w_s = r \cdot w_b, \quad h_s = w_s \cdot h_f/w_f,
\end{equation}

Instead of annotating the objects' bounding boxes directly, which is the standard practice in visual grounding and object detection, we convert bounding boxes into coordinates to make them better suited for use with MLLMs. Given the center position $(x, y)$, the bounding box $(a, b, c, d)$ representing left, bottom, right, and top edges is calculated as:

\vspace{-0.2cm}
\begin{equation}
\begin{aligned}
a &= x \cdot w_b - \frac{1}{2} w_t, &\quad b &= y \cdot h_b - \frac{1}{2} h_t \\
c &= x \cdot w_b + \frac{1}{2} w_t, &\quad d &= y \cdot h_b + \frac{1}{2} h_t
\end{aligned}
\end{equation}

The continuous action space is discretized into two subspaces: \ding{182} \textit{Movement Action Subspace:} MoveUp, MoveDown, MoveLeft, and MoveRight. \ding{183} \textit{Resizing Action Subspace:} Enlarge and Reduce.


Because MLLMs struggle to accurately predict appropriate step sizes, we manually set the initial step sizes for both movement and resizing. The optimization problem is highly non-convex, with many sub-optimal placement configurations. The objective function is mostly flat (i.e., most placements are irrational), with small, sensitive regions where slight changes matter. To explore this complex space effectively, we use a wide range of initial step sizes, such as 0.05 to 0.95 for movement and 0.3 to 3.0 for resizing.

Each action is symbolically represented by a unique alphabetic encoding for clarity when prompting MLLM. At each step, the MLLM selects an action based on the current state, and the corresponding operation is executed using the current step size. Additionally, we introduce void actions, “No movement needed” and “No resizing needed”, which allow the MLLM to indicate convergence.

\presto~achieves significantly greater compactness than standard search algorithms, as it uses an MLLM to deterministically choose a single movement out of four and a resizing action out of two at each step. This reduces the search space by 87.5\%. Our ablation study (Section \ref{subsec:ablation}) shows that MLLM enables better convergence given an iteration budget.

\subsection{Selection}

The selection stage in \presto~is designed to identify placements with the highest semantic coherence. Our \presto~framework uses both predefined metrics and MLLM to guide the selection: \ding{182}\textbf{Metric-guided Selection:} This strategy is similar to fitness scoring in genetic algorithms. A predefined metric is used to rank placement candidates. In our \presto~framework, we use SimOPA, which takes a synthetic composite image and its corresponding mask as input and returns a score indicating the plausibility of the placement. \ding{183} \textbf{MLLM-as-a-judge:} To leverage the MLLM’s built-in visual and commonsense knowledge, we let it directly assess object placements. The MLLM compares synthesized image pairs and selects the one with the more reasonable placement. Since placement quality can be subjective and influenced by many factors, we prompt the MLLM to focus specifically on the foreground object’s position and scale.

    

For MLLM-as-a-judge, we implement a tournament-style selection process: candidates are compared in sequential pairs based on their initial ordering, with the less plausible option eliminated in each round. This process continues until a single best placement remains. Full details are provided in Appendix C.


Although SimOPA is the most widely adopted metric for evaluating placement, we find that MLLM-based judgments align more closely with human perception. A detailed comparison of the two strategies is presented in Section \ref{subsec:quality}.


\section{Experiment}

\subsection{Settings and Metrics}\label{subsec:details}

We evaluate \presto~using GPT-4o as the backbone MLLM~\cite{hurst2024gpt4o}. During initialization, we set the number of initial candidates $k = 4$ and use a default scaling ratio of 0.2. In the iterative heuristic optimization phase, we use a decay factor $\alpha = 0.7$ and limit the maximum number of iterations to $T = 3$. More details on hyperparameter choices and ablation analysis can be found in Section \ref{subsec:ablation}.

We follow prior work and use SimOPA accuracy and FID~\cite{heusel2017gans} to measure placement credibility, and LPIPS~\cite{zhang2018unreasonable} to measure placement diversity. We conduct experiments on the OPA dataset~\cite{liu2021opa}, which consists of 73,470 images collected from MS COCO~\cite{lin2014microsoft}, with 1,389 backgrounds, and 4,137 foreground objects across 47 categories. Additionally, we test on the OPAZ dataset~\cite{qin2025think}, which includes novel objects and scenes not present in OPA, for open-world evaluation. Detailed descriptions are illustrated in Appendix~A.

\begin{table*}[ht]
    \vspace{-0.5cm}
    \centering
    \renewcommand{\arraystretch}{0.94}
    \small
    \caption{Comparison results on OPA and OPAZ datasets. \presto-Metric: \presto-with Metric-guided Selection. \presto-MLLM: \presto~with MLLM-as-a-judge. Hum: Average rationality ratings (out of 10) of the placements by survey participants.}
    \label{tab:results}
    \begin{tabular}{p{20mm}<{\centering}|p{13mm}<{\centering}p{13mm}<{\centering}p{13mm}<{\centering}|p{13mm}<{\centering}p{13mm}<{\centering}p{13mm}<{\centering}|p{15mm}<{\centering}}
        \hline
        \multirow{2}{*}{Method} & \multicolumn{3}{c|}{OPA} & \multicolumn{3}{c|}{OPAZ} & \multirow{2}{*}{LPIPS$\uparrow$} \\
        \cline{2-7}
        & Hum $\uparrow$ & Acc$\uparrow$ & FID$\downarrow$ & Hum $\uparrow$ & Acc$\uparrow$ & FID$\downarrow$ & \\
        \hline
        TERSE~\cite{tripathi2019learning} & 3.86 & 0.679 & 46.94 & 3.11 & 0.340 & 81.1 & 0 \\
        PlaceNet~\cite{zhang2020learninga} & 3.43 & 0.683 & 36.69 & 3.90 & 0.367 & 63.6 & 0.160 \\
        GracoNet~\cite{zhou2022learning} & 4.54 & 0.847 & 27.75 & 3.66 & 0.431 & 59.1 & 0.206 \\
        CA-GAN~\cite{wang2023cagan} & 3.79 & 0.792 & 23.21 & 3.74 & 0.347 & 49.0 & 0.268 \\
        CSANet~\cite{wang2024learning} & 4.10 & 0.863 & 20.88 & 4.24 & 0.426 & 45.1 & \textbf{0.274} \\
        IOPRE~\cite{zhang2023interactive} & 4.42 & 0.895 & 21.59 & 3.63 & 0.586 & \textbf{27.8} & 0.214 \\
        CSENet~\cite{qin2025think} & - & 0.940 & \textbf{17.51} & - & 0.618 & 42.1 & 0.137 \\
        \hline
        \presto-Metric & 7.09 & \textbf{0.958} & 23.05 & 5.97 & \textbf{0.757} & 35.9 & 0.171 \\
        \presto-MLLM & \textbf{7.09} & 0.926 & 19.79 & \textbf{7.20} & 0.620 & 57.1 & 0.119 \\
        \hline
    \end{tabular}
    \vspace{-0.5cm}
\end{table*}

\begin{figure*}[t]
  \centering
  \includegraphics[width=\linewidth]{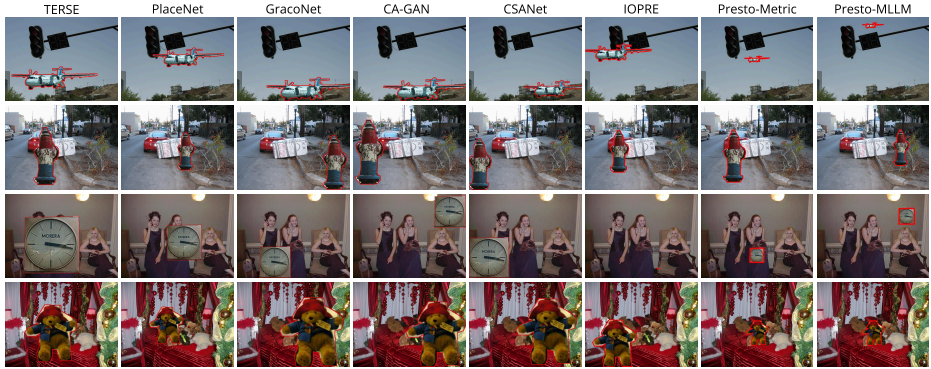}
  \caption{Example placements by \presto~and the state-of-the-art methods. The foreground objects are outlined in red.}
  \label{fig:vis0}
  \vspace{-0.5cm}
\end{figure*}

\subsection{Main Results}
We compare \presto~with seven state-of-the-art object placement techniques that are detailed in Appendix~B. As shown in Table \ref{tab:opa}, on the OPA dataset, \presto~with Metric-guided Selection achieves the highest accuracy of 0.958, outperforming all baselines. Even without optimizing for SimOPA, \presto~with MLLM-as-a-judge achieves a strong accuracy of 0.926, surpassing six of the seven baselines. It is important to note that all baselines are supervised models trained directly on the OPA dataset, leveraging prior knowledge of object placement within similar domains. In contrast, the MLLM backbone has never seen the OPA dataset during training, and \presto~operates in a zero-shot, training-free setting. Furthermore, \presto~is efficient with fast convergence: the Metric-guided variant converges in just 1.02 iterations on average, while the MLLM-as-a-judge variant converges in 2.61 iterations.

In terms of FID, \presto~also performs well. With Metric-guided Selection, it achieves a score of 23.05, outperforming four baselines. The MLLM-as-a-judge variant performs even better, beating six baselines. These results suggest that \presto~generates visually realistic placements. Also, while \presto~is outperformed by CSENet in FID, we demonstrate in Section \ref{subsec:quality} that human evaluators find \presto’s placements more realistic.

For the OPAZ dataset, the overall accuracy drops due to the increased difficulty, as expected. Nevertheless, both variants of \presto~outperform all baselines in accuracy. Notably, Metric-guided Selection achieves a 23.9\% to 41.7\% accuracy improvement, demonstrating \presto’s strong generalization to open-world object placement scenarios. In terms of FID, Metric-guided \presto~outperforms six of seven baselines, trailing only IOPRE. However, note that it achieves a 17.1\% higher accuracy than IOPRE.

\presto~achieves satisfactory diversity, although its LPIPS scores are lower than those of some state-of-the-art baselines. Interestingly, we observe that \presto’s placements remain stable across different temperature settings, suggesting that the MLLM leverages its understanding of the visual context to make consistent, well-informed decisions without introducing random variation. The MLLM-as-a-judge strategy yields an even lower LPIPS score of 0.119, further indicating that its perceptual judgments are more stable than those driven by the SimOPA when evaluating similar placement configurations. Methods like CA-GAN and CSANet achieve higher LPIPS scores, but often at the expense of accuracy. In contrast, \presto~achieves a better balance: it maintains reasonable diversity while avoiding the uncontrolled variability which leads to unrealistic or irrational placements. Notably, \presto~outperforms CSENet, the most accurate baseline, in both accuracy and diversity.

\begin{figure}[t]
  \centering
  \includegraphics[width=\linewidth]{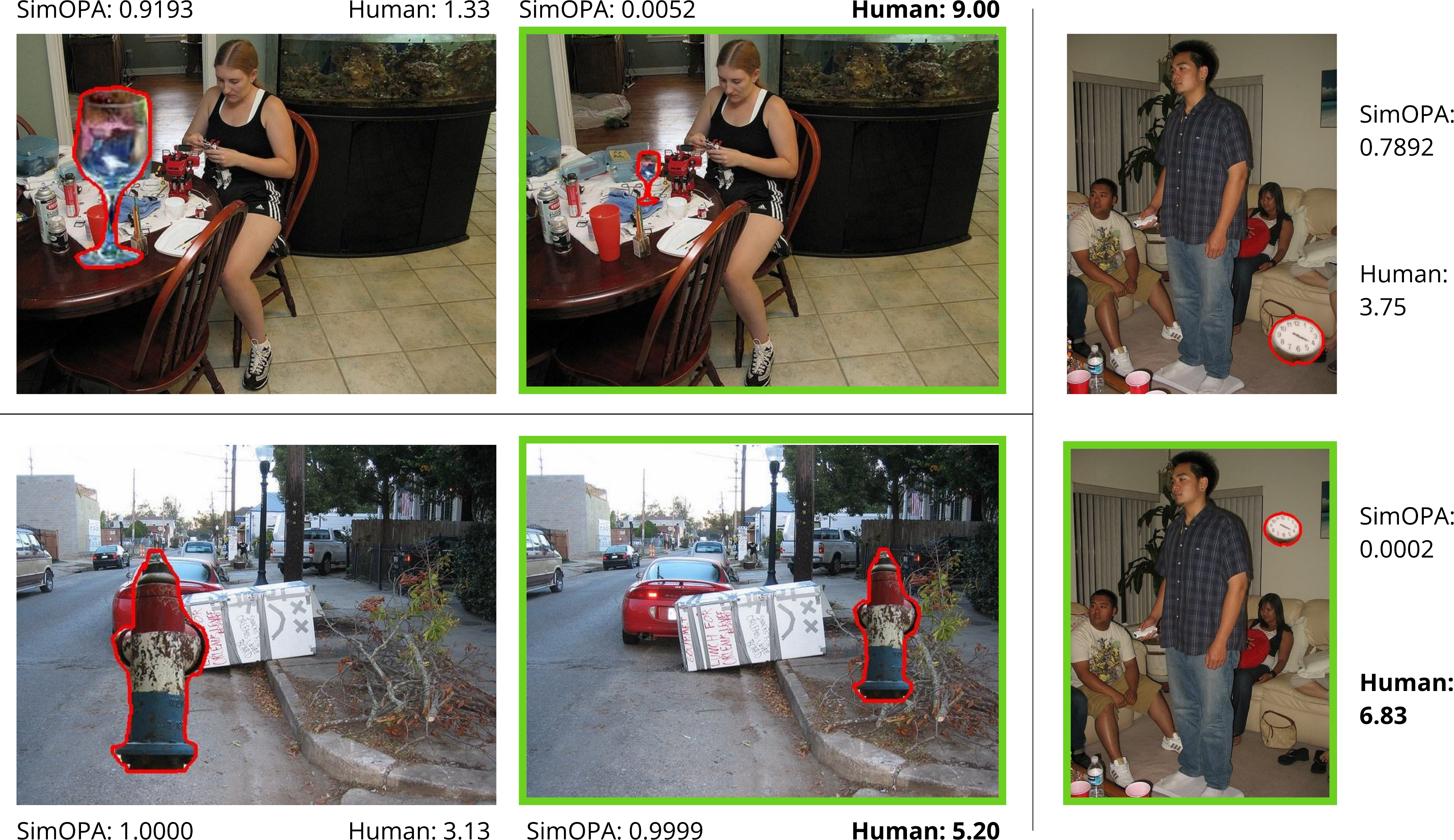}
  \caption{Comparison between OPA scores and human ratings. Selections made by the MLLM are outlined in green.}
  \label{fig:relevance}
  \vspace{-0.5cm}
\end{figure}



\subsection{Qualitative and Human Evaluation} \label{subsec:quality}

Figure \ref{fig:vis0} presents visual comparisons between object placements generated by \presto~and six state-of-the-art baselines. CSENet is excluded due to the unavailability of their model checkpoint. Across examples, \presto~consistently produces more plausible placements in terms of object location and scale. We attribute this improvement to two key factors: (1) the incorporation of common sense reasoning via MLLMs, which prevents clearly illogical placements (e.g., clocks are mounted on walls rather than on people; fire hydrants are placed near sidewalks rather than in the middle of roads); and (2) the MLLM’s understanding of spatial context, especially relative scale (e.g., airplanes appear smaller than nearby traffic lights due to distance; the teddy bear is appropriately scaled relative to other toys on the bed). More examples are provided in Appendix~D.


\begin{table}[htbp]
    \vspace{-0.5cm}
    \centering
    \small
    \caption{Ablation study on the role of MLLM.}
    \label{tab:as}
    \setlength{\tabcolsep}{4pt}
    \begin{tabular}{c|c|c c c c}
        \hline
        & Metric & w/o MLLM & \begin{tabular}{@{}c@{}}optimization\\only\end{tabular} & \begin{tabular}{@{}c@{}}initialization\\only\end{tabular} & \presto-Metric \\
        \hline
        \multirow{2}{*}{\emph{Credibility}} & acc.$\uparrow$ & 0.79 & 0.93 & 0.90 & \textbf{0.99} \\
        & FID$\downarrow$ & 89.1 & 83.7 & 82.9 & \textbf{80.5} \\
        \hline
        \emph{Diversity} & LPIPS$\uparrow$ & 0.136 & 0.043 & 0.154 & \textbf{0.167} \\
        \hline
    \end{tabular}
    \vspace{-0.5cm}
\end{table}

To validate these observations, we conducted an anonymous user study using an online questionnaire. Each participant evaluated 15 composite images randomly drawn from the OPA or OPAZ datasets. For each image, three versions were shown: (1) the original background, (2) the composite with the inserted object highlighted in red, and (3) the composite without the outline. Participants rated the rationality of the object placement on a 10-point scale. A rubric guides them to focus on depth, relative size, occlusion, physical support, and semantic appropriateness, and ignore factors such as resolution, lighting, and color matching. The questionnaire design is presented in Appendix~F.

The study was approved by the Institutional Review Board (IRB) at our University. We collected 1,320 ratings from 88 responses from faculty and students. As shown in Table \ref{tab:results}, placements generated by \presto~received significantly higher ratings than those from all baseline methods. On both OPA and OPAZ, the baselines averaged between 3.11 and 4.54, with slightly lower scores on OPAZ. In contrast, \presto~achieved average ratings of 7.09 on OPA and 5.97 and 7.2 on OPAZ.

Notably, the MLLM-as-a-judge variant received higher human ratings than the Metric-guided Selection strategy, despite achieving lower accuracy. This discrepancy suggests a misalignment between SimOPA and human perceptual judgments. As shown in Figure \ref{fig:relevance}, certain placements that received high SimOPA scores were rated as irrational by users, while placements generated by the MLLM-as-a-judge variant were consistently rated as more reasonable. These results support our hypothesis: MLLMs not only guide object placement more rationally than existing methods, but also provide more perceptually aligned evaluations than existing metrics.

\subsection{Ablation Study} \label{subsec:ablation}

\begin{table}[ht]
    \vspace{-0.5cm}
    \centering
    \small
    \caption{Ablation study on choices of initial candidates number $k$, initial scaling factor $\mathbf{r^{(0)}}$, and decay factor $\alpha$.}
    \label{tab:ha}
    \setlength{\tabcolsep}{3pt}
    \resizebox{\linewidth}{!}{%
        \begin{tabular}{c|c|ccc|ccc|ccc}
        \hline
        & \multirow{2}{*}{Metric} & \multicolumn{3}{c|}{$k$} & \multicolumn{3}{c|}{$\mathbf{r}^{(0)}$} & \multicolumn{3}{c}{$\alpha$} \\
        \cline{3-11}
        & & 1 & 4 & 8 & 0.1 & 0.2 & 0.5 & 0.3 & 0.7 & 1.0 \\
        \hline
        \multirow{2}{*}{\emph{Credibility}} & acc.$\uparrow$ & 0.94 & \textbf{0.99} & 0.93 & 0.51 & \textbf{0.99} & 0.78 & 0.95 & \textbf{0.99} & 0.92 \\
        & FID$\downarrow$ & 86.8 & \textbf{80.5} & 89.2 & 82.0 & 80.5 & \textbf{74.5} & \textbf{80.3} & 80.5 & 81.1 \\
        \hline
        \emph{Diversity} & LPIPS$\uparrow$ & 0.138 & 0.167 & \textbf{0.194} & 0.105 & \textbf{0.167} & 0.152 & 0.151 & 0.167 & \textbf{0.181} \\
        \hline
        \end{tabular}
    }
    \vspace{-0.5cm}
\end{table}

We evaluate the effect of three key hyperparameters: the number of initial candidates $k$, the initial scaling factor $\mathbf{r}^{(0)}$, and the decay factor $\alpha$. Experiments are conducted on 100 randomly selected image pairs from the OPA dataset. As shown in Table \ref{tab:ha}, setting k = 4 achieves the best balance. Smaller values reduce the diversity of initial placements, while larger values increase noise in MLLM predictions, leading to less accurate results. For the initial scaling factor, a moderate value of 0.2 outperforms both smaller (0.1) and larger (0.5) settings, providing better flexibility for size adjustment. Finally, we test decay factors $\alpha = 0.3, 0.7, 1.0$ and observe that while performance is generally stable, a larger $\alpha$ tends to overly restrict the search space, causing early convergence to suboptimal placements.

                    
                     
                     
                        
                        
                        
                       

We further analyze the role of the MLLM by disentangling its contributions to initialization and iterative heuristic optimization. The maximum number of iterations is set to 3. We first remove MLLM entirely, replacing both stages with random actions. This results in a performance drop of over 20\%, underscoring the importance of MLLM guidance. Next, we isolate each component to evaluate its individual contribution. When the MLLM is used only for optimization, accuracy improves by 14\% over the baseline. When used only for initialization, the improvement is 18\%. 



\begin{table}[ht]
    \centering
    \small
    \caption{Ablation study on the generalizability across MLLMs.}
    \label{tab:general}
    \setlength{\tabcolsep}{3pt}
    \begin{tabular}{c|c|c c|c c}
        \hline
        Metric & Dataset & Qwen-VL2.5 & DeepSeek-VL2 & \begin{tabular}{@{}c@{}}Qwen-VL2.5\\+\presto-Metric\end{tabular} & \begin{tabular}{@{}c@{}}DeepSeek-VL2\\+\presto-Metric\end{tabular} \\
        \hline
        \multirow{2}{*}{Accuracy $\uparrow$} & OPA & 0.34 & 0.63 & 0.85 & 0.91 \\
        & OPAZ & 0.32 & 0.38 & 0.61 & 0.71 \\
        \hline
    \end{tabular}
    \vspace{-0.5cm}
\end{table}

We also apply \presto~to more affordable, open-source MLLMs such as Qwen-VL2.5-7B~\cite{Qwen2.5-VL} and DeepSeek-VL2-27B~\cite{wu2024deepseekvl2}. As shown in Table \ref{tab:general}, the results highlight \presto’s strong generalization across different model sizes. Qwen-VL2.5 achieves SimOPA accuracies of 85\% and 61\%, outperforming four and six state-of-the-art methods, respectively. Moreover, \presto~consistently improves performance across all models, boosting accuracy by 51\% and 29\% for Qwen-VL2.5-7B, and by 28\% and 33\% for DeepSeek-VL2-27B. 

\vspace{-0.4cm}
\section{Conclusion}\label{sec:con}
In this work, we tackle the limitations of current object placement methods that rely on heavy training and annotated data. We reformulate the task as a heuristic search problem and propose \presto, a novel, zero-shot, training-free framework. \presto~uses MLLM-guided decisions within an imaginary search space to iteratively adjust an object’s position and scale. Experiments on OPA and OPAZ datasets show that \presto~achieves outstanding results without category-specific supervision, outperforming baselines on both standard metrics and human alignment. Its plug-and-play design and strong generalization make it suitable for open-world applications.



\bibliographystyle{splncs04}
\bibliography{iconip26.bib}

\end{document}